\documentclass{ecai}

\usepackage{times}
\usepackage{graphicx}
\usepackage{latexsym}
\usepackage{helvet} 
\usepackage{courier} 
\usepackage[hyphens]{url}  
\usepackage{subcaption}
\usepackage{amsfonts}
\usepackage{booktabs}
\usepackage{tabularx}
\usepackage{amsmath} 
\usepackage{subcaption}
\usepackage[skip=0pt]{caption}
\usepackage{multirow}
\usepackage[inline]{enumitem}
\usepackage{acronym}
\usepackage{color}
\usepackage{colortbl}
\usepackage{cite}
\usepackage[numbers]{natbib}
\usepackage{enumitem}
\usepackage[dvipsnames]{xcolor}

\newcommand{\shrink}[1]{\vspace*{-2mm}}

\acrodef{STR}{Scene Text Recognition}
\acrodef{Bi-STET}{Bidirectional Scene Text Transformer}
\acrodef{RNN}{Recurrent Neural Network}
\acrodef{CNN}{Convolutional Neural Network}
\acrodef{STN}{Spatial Transformer Network}
\acrodef{STET}{Scene Text Transformer}
\acrodef{FAN}{Focusing Attention Network}
\acrodef{VFE}{Visual Feature Embedding}
\acrodef{VFEN}{Visual Feature Extraction Network} 
\acrodef{CTC}{Connectionist Temporal Classification}
\acrodef{SOTA}{state-of-the-art} 
\def\BibTeX{{\rm B\kern-.05em{\sc i\kern-.025em b}\kern-.08emT\kern-.1667em\lower.7ex\hbox{E}\kern-.125emX}}
\DeclareMathOperator*{\softmax}{softmax}
\DeclareMathOperator*{\Attention}{Attention}

\ecaisubmission   

\begin{document}

\title{Bidirectional Scene Text Recognition \\with a Single Decoder}

\author{Maurits Bleeker\institute{University of Amsterdam,
		The Netherlands, email: m.j.r.bleeker@uva.nl}  
		\and Maarten de Rijke\institute{University of Amsterdam,
		The Netherlands, email: derijke@uva.nl}}

\maketitle

\begin{abstract}
	\ac{STR} is the problem of recognizing the correct word or character sequence in a cropped word image. 
	To obtain more robust output sequences, the notion of bidirectional \ac{STR} has been introduced.
	So far, bidirectional \acp{STR} have been implemented by using two separate decoders; one for left-to-right decoding and one for right-to-left. 
	Having two separate decoders for almost the same task with the same output space is undesirable from a computational and optimization point of view.
	We introduce the \ac{Bi-STET}, a novel bidirectional \ac{STR} method with a single decoder for bidirectional text decoding. 
	With its single decoder, \ac{Bi-STET} outperforms methods that apply bidirectional decoding by using two separate decoders while also being more efficient than those methods, 
	Furthermore, we achieve or beat \ac{SOTA} methods on all \ac{STR} benchmarks with \ac{Bi-STET}. 
	Finally, we provide analyzes and insights into the performance of \ac{Bi-STET}.
\end{abstract}

\section{INTRODUCTION}

\acf{STR} is the task of recognizing the correct word or character sequence in a cropped word image. 
Many different architectures have been proposed for \ac{STR}.   
Since the rise of deep learning, most state-of-the-art \ac{STR} methods adopt a \acf{CNN} for feature extraction and an encoder-decoder  architecture as the core component for sequence modeling. After feature extraction with a \ac{CNN}, the extracted features of the input image are encoded into a new representation with an encoder. As a final step, conditioned on the encoded input image representation,  the character sequence is decoded, which is depicted in the input image.
Figure~\ref{fig:model-overview} summarizes a general pipeline.
\citet{baek2019wrong} identify sequence modeling as a core component in \ac{STR} frameworks.

The encoder-decoder architecture for the sequence modeling stage of many state-of-the-art \ac{STR} methods (see Section~\ref{sec:related}) can be characterized by 
\begin{enumerate*}[label={(\roman*)}]
\item a bidirectional \ac{RNN}  for feature encoding,
\item a directed \ac{RNN} decoder  for character decoding, and
\item various types of attention mechanism~\citep{bahdanau2014neural, luong2015effective} to generate additional context vectors for the current decoding steps.
\end{enumerate*}

Two recent developments have accelerated progress in \ac{STR}: 
\begin{enumerate*}[label={(\roman*)}]
\item a move away from recurrent sequence modeling, and 
\item bidirectional decoding for \ac{STR}.
\end{enumerate*}

Regarding the first, \citet{sheng2018nrtr} have changed the standard \ac{STR} approach for sequence modeling by introducing a non-recurrent method based on a transformer encoder-decoder architecture~\citep{vaswani2017attention}. By using a transformer-based encoder-decoder, the model architecture can be simplified and the time for model optimization can be reduced by an order of magnitude in comparison~\citep{sheng2018nrtr}. 

The second reason for recent progress in \ac{STR} is  bidirectional decoding. Bidirectional decoding is the idea of decoding an output sequence in two directions (i.e., from left-to-right and right-to-left) for more robust output predictions. 
This bidirectional decoding is implemented by using a different decoder for each decoding direction \cite{shi2018aster}. Decoding the text in two directions at the same time can be seen as two different sub-tasks for the model to perform.

It is important to reflect on different ways of modeling sub-tasks, especially using task conditioning.
With task conditioning, the output of a method does not solely depend on the input data, but on a given (sub-)task as well. 
In other words, given the same input data, the output may be different based on the (sub-)task it is conditioned on. 
As explained by \citet{radford2019language}, task conditioning can be implemented in several ways. 

One option is at the \emph{algorithmic} level \citep{finn2017model}, where different models are learned for different tasks, and an overall algorithm selects the correct model for a particular task. 
Implementing task conditioning at the algorithmic level is not optimal, since in most cases, there is a lot of shared knowledge between different (sub-)tasks, which is not exploited when separate models are optimized for each task.
Another way of implementing task conditioning is at the \emph{architecture} level.  
For bidirectional \ac{STR}, \citet{shi2018aster} have implemented the decoding direction as two sub-tasks at the architecture level, that is, by having two separate decoders: one for left-to-right and another one for right-to-left text decoding. 
Although both decoders share the same encoder, two separate decoders are optimized for two tasks that are (almost) identical and share the same output space. 

Implementing bidirectional decoding at the architecture level is not uncommon, and has been done for other tasks besides \ac{STR}~\citep{zhang2018asynchronous, zhou2019synchronous}.
However, having two separate decoders for two tasks that are similar (i.e., left-to-right and right-to-left \ac{STR}) is not desirable: 
\begin{enumerate}[label={(\arabic*)},leftmargin=*,nosep]
\item From a computational point of view: The two network components do not share weights, which requires separate optimization for both parts.
\item From a multi-task learning point of view: There is a lot of shared knowledge between left-to-right and right-to-left decoding and both tasks share the same output space, which is not utilized when optimizing the two decoders apart from each other. 
\end{enumerate}
Therefore, the question remains:  \emph{Can we have the benefits of bidirectional decoding (left-to-right and right-to-left decoding) for \ac{STR} without implementing this at the architecture or algorithmic level?}

There is promising room for improvement on the implementation side of bidirectional decoding for \ac{STR} by just using one decoder for both decoding directions. Instead of implementing this decoding direction at the algorithmic or at the architecture level, we propose a new way of implementing task conditioning, namely, at the \emph{input} level. Implementing task conditioning at the input level means that extra feature information is added to the input of the model. This context information should be exploited by the model so as to condition on the right (sub-)task.

More specifically, the transformer architecture, as used by \citet{sheng2018nrtr} for the encoder-decoder part for \ac{STR}, has no recurrent inductive bias. 
To solve sequential problems with transformers at the input level, additional position embeddings are added to provide the model with information about the order of the input sequence. 
Due to the ``position unawareness'' of the transformer, the model is also not limited to an inductive decoding direction (unlike RNNs). 
By adding an extra embedding to the input data, which tells the method to decode an input example from left-to-right or right-to-left, the model can exploit bidirectional decoding with one unified architecture for both directions. 
This means that the model has one decoder with one set of model parameters that can be optimized for both subtasks at the same time. 
This is in contrast with the method by \citet{shi2018aster}, where two decoders are optimized, one for each decoding direction. 

In this work, we show that we can simplify the bidirectional \ac{STR} architecture by using a transformer based encoder-decoder which is able to perform bidirectional text recognition by using a single decoder. 
Our main technical contributions in this paper are the following:
\begin{itemize}[nosep,leftmargin=*]
\item We introduce \acs{Bi-STET}\acused{Bi-STET},  \textbf{BI}directional \textbf{S}cene \textbf{TE}xt \textbf{T}ransformer. 
\ac{Bi-STET} is a unified network architecture, optimized for two sub-tasks (left-to-right and right-to-left \ac{STR}), using one forward pass. 
We achieve this through the implementation of bidirectional decoding at the input level as opposed to previous works which do this at the architecture level~\cite{shi2018aster}.
We condition the output sequence on a specific decoding direction by adding extra features at the input level, which results in a direction-agnostic decoder architecture.   
It is possible to exploit the transformer architecture for task conditioning at the input level, without requiring additional model components or algorithms to model this task conditioning. 
\item We show that \ac{Bi-STET} achieves or outperforms state-of-the-art \ac{STR} methods with a simpler and more efficient approach than other bidirectional \ac{STR} methods. We achieve these similar results with fewer weight parameters and 50\% less training iterations. 
\item We provide analyzes and insights on the performance of \ac{Bi-STET}.\footnote{For reproducibility and repeatability, the code and checkpoint files used to train and evaluate \ac{Bi-STET} will be made available at \url{https://github.com/MauritsBleeker/Bi-STET}.} We analyze the generalisation of  \ac{Bi-STET} w.r.t. oriented and curved text, the learned attention mechanism of the encoder-decoder, the relation between sequence length and test-accuracy of \ac{Bi-STET} and is \ac{Bi-STET} makes more mistakes for rarer expressions than with dictionary words. 
\end{itemize}

\section{RELATED WORK}\label{sec:related}

Traditional methods for \ac{STR}~\citep{bissacco2013photoocr, shivakumara2011new} apply a  bottom-up approach.  
The input image is preprocessed for feature extraction and character segmentation is applied to obtain single characters from the input image for word inference. 
For an overview, see \citep{zhu2016scene}.  With the rise of deep learning, \ac{STR} methods increasingly focus on end-to-end training from the input image to the desired output character sequence.

\subsection{Deep-learning based text recognition}

\citet{jaderberg2014deep} have proposed the first method for unconstrained \ac{STR} with deep learning.  
The method predicts a sequence of characters with a fixed length by using a \ac{CNN} classification model.   
A bag-of-N-grams is also predicted; representing an unordered set of character N-grams that occur in the word depicted in the input image. 
The predictions are combined in a Path Select Layer to predict the most likely character sequence.  
More recently, \citet{jaderberg2016reading} propose another method where the recognition task is formulated as a multi-class classification problem over a 90k-class lexicon.

Many other \ac{STR} methods \citep{zhan2018esir, shi2016robust, shi2017end, shi2018aster, cheng2017focusing, yang2019symmetry}  use a \ac{CNN} for feature extraction in combination with an encoder-decoder model to map the sequence of image features to a character sequence. To solve the problem of rotation invariance of \ac{CNN}s (i.e., for curved and oriented text), several methods have been proposed \citep{shi2018aster, shi2016robust, zhan2018esir, yang2017learning, yang2019symmetry}. 
\citet{shi2018aster, shi2016robust} and \citet{yang2019symmetry} use a \acf{STN} for input image rectification to handle perspective text and curved text.  
\citet{zhan2018esir} have introduced an iterative Rectification Network where the input image is rectified multiple times by removing perspective distortion and text line curvature.

The alignment between the predicted character and the corresponding region in the input image is modelled with two different approaches. 
The first approach is to use CTC loss~\citep{shi2017end, su2014accurate, liu2016star, gao2017reading, graves2009novel}. 
The other is to connect the \ac{RNN} encoder to the decoder via an attention mechanism \citep{shi2018aster, zhan2018esir, shi2016robust, cheng2017focusing, yang2017learning}, which creates an additional context vector for the encoded input image conditioned on the already predicted output sequence.

\citet{shi2018aster} introduce the notion of bidirectional \ac{STR}. 
Each output sequence is predicted in two directions with two separate decoders, which do not share parameters. 
The output sequence with the highest probability is selected as the final prediction in order to obtain more robust predictions.  
\citet{sheng2018nrtr} are the first to use a non-recurrent encoder-decoder approach based on a transformer architecture; they have also introduced a modality-transform block to map an image to a sequence feature representation.

In contrast to most work in \ac{STR}, we do not use any specific component for image rectification. We also do not rely on \ac{RNN}s for sequence modeling. 
Similar to \citet{sheng2018nrtr}, we also use a transformer architecture which yields state-of-the-art results in text recognition without using extra image rectification components, for bidirectional sequence modelling  \ac{STR}. 
However, we achieve this by using a single decoder for the bidirectional decoding, resulting in significantly fewer parameters and training iterations. 

\begin{figure*}
    \centering
	\includegraphics[width=.55\textwidth]{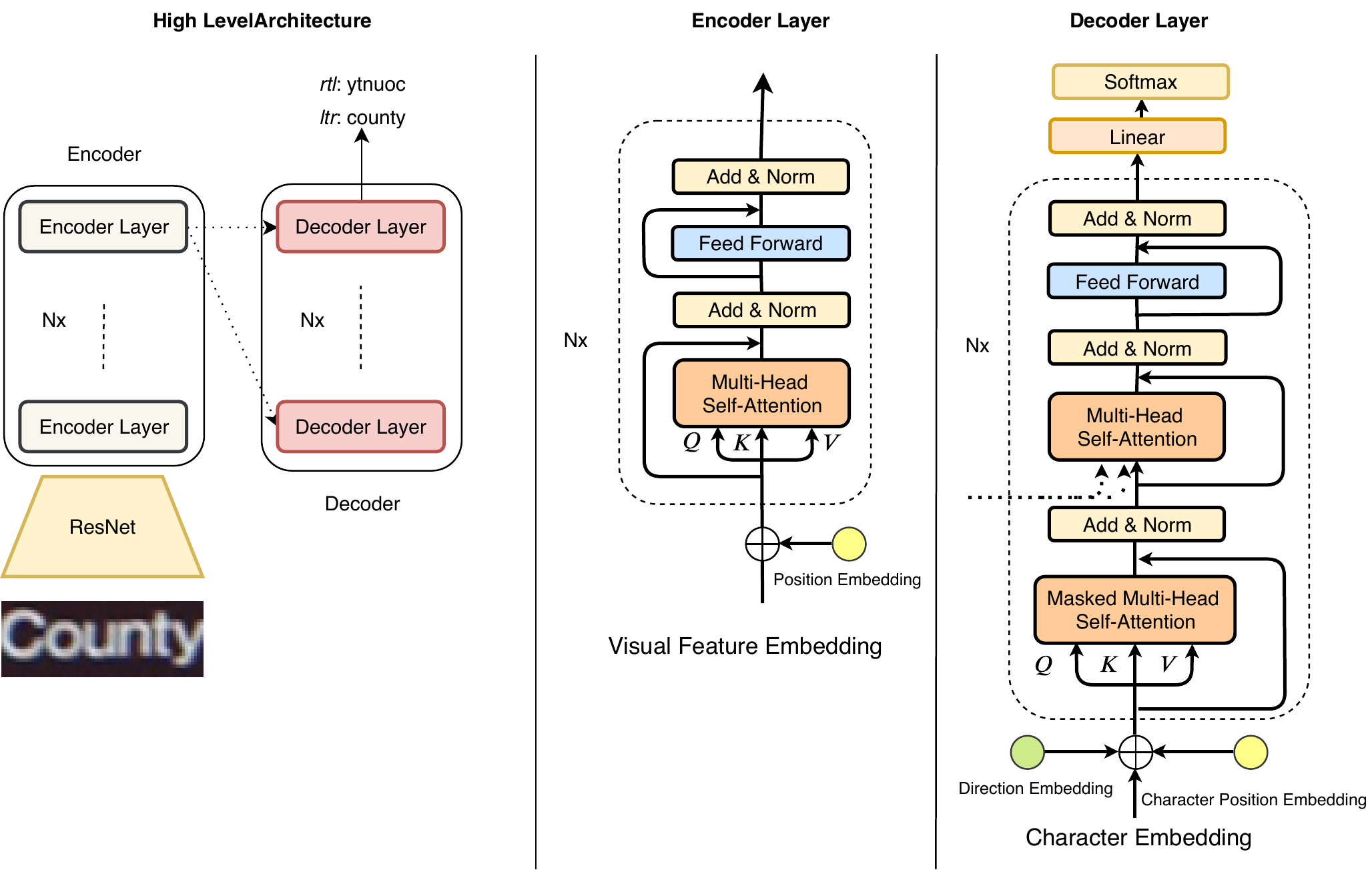}
	\caption{Overview of \ac{Bi-STET}. A ResNet architecture is used for visual feature extraction. Next, a stack of $n$ transformer encoder layers is used for encoding the visual image features. For decoding the output sequence, a stack of $n$ decoder transformer layers is used.}
	\label{fig:model-overview}
\end{figure*}

\subsection{Task conditioning}

As indicated by \citet{radford2019language}, the distribution over the possible model outputs is naturally modelled as $p(y\mid x)$, where $x$ is the input data, and $y$ is a possible output prediction. 
With \emph{task conditioning} (or modality conditioning), the output is not only conditioned on the input data, but also on a given task, dataset, or modality as well, i.e., $p(y\mid x, t)$, where $t$ stands for the \textit{task}.

One way of implementing task conditioning is at the \textit{architecture} level.  
\citet{kaiser2017one} introduce a single model for eight tasks; for each data-modality and/or subtask, different encoders and decoders are used with a shared latent space. 
\citet{devlin2018bert} adopt the transformer to train a task-agnostic language model. 
After training the language model, additional task-specific output layers are fined-tuned for each evaluation task.  
\citet{finn2017model} introduce a meta-learning framework for multi-task learning where  task conditioning is implemented at the \textit{algorithmic} level: tasks are treated as training examples and are sampled from a distribution over tasks during training.
During each training iteration, for each task, a separate model for each task is updated based on the loss for that specific task.

Unlike previous work, we condition a subtask (i.e., the decoding direction) at the input level. 
As a result, we do not need different models or network components for each decoding direction. 
Having only one decoder is desirable from both a computational point of view (i.e., only one decoder to optimize) and from an optimization point of view (i.e., shared weights for all sub-tasks).


\section{METHOD}
\label{section:method}

To address the \ac{STR} task, we take a fixed size image $\mathcal{I}$ as input and want to decode the sequence of output characters $y_1, \dots, y_{L}$, where $L$ is the length of the character sequence depicted in the input image.  
Briefly, we use a multi-layer stack of transformers for both the encoder and decoder.  
We use exactly the same implementation of the transformer as described in \citep{vaswani2017attention} for the encoder-decoder. 
Therefore, we refer to \citep{vaswani2017attention} for the exact details of the implementation.
A full overview of \ac{Bi-STET} is shown in Figure~\ref{fig:model-overview}.

\subsection{Visual feature extraction network}

Like \citep{shi2018aster, zhan2018esir, cheng2017focusing, yang2019symmetry}, we use a ResNet \citep{he2016deep} based architecture for the \ac{VFEN}. 
A  ResNet architecture is a more suitable feature extractor than VGG \citep{simonyan2014very} for STR, as shown in \citep{shi2018aster, zhan2018esir}.
We use a 45-layer residual network, with the same network configuration as \citep{shi2018aster}. 
We split the obtained feature representation $\mathcal{Q}  \in \mathbb{R}^{W \times C \times H}$ column-wise, which results in a sequence of \acfp{VFE}, $ \textbf{v}_{1},\ldots,\textbf{v}_{W}$, where $ \textbf{v}_{i} \in \mathbb{R}^{C \times H}$. 
 
\subsection{Feature encoding}
\label{sec:encoder}

The feature encoder is in charge of encoding the visual image embeddings. 
Each visual image embedding is encoded into a new representation in $n$ steps, by using transformer encoder layers, while attending over the entire sequence of \ac{VFE}s during each encoding step. 
We use  scaled dot-product as the attention function:
\begin{equation}
\Attention(\textbf{Q}, \textbf{K}, \textbf{V}) = \softmax\left( \frac{\textbf{Q}\textbf{K}^{T}}{\sqrt{d}}\right)\textbf{V}.
\end{equation}
Scaled dot-product attention can be described as a weighted sum of the vectors in matrix $\textbf{V}$, which is a horizontal concatenation of the flattened sequence of \acp{VFE} (also referred to as \textit{values}). 
Each embedding $\textbf{v}$ is weighted by the similarity between a key $\textbf{k}$ and a query $ \textbf{q}$.  
In each transformer layer multiple heads of attention are used: 
\begin{equation}
\mathit{head}_{i} = \Attention\left(\textbf{Q} \textbf{W}^{Q}_{i}, \textbf{K} \textbf{W}^{K}_{i}, \textbf{V} \textbf{W}^{V}_{i}\right) 
\end{equation}
and
\begin{equation}
\begin{split}
\mbox{}\hspace*{-2.5mm}
\mathit{MultiHeadSelfAttention} \,{=}\,  \mathit{Con}\mathit{Cat}(&\mathit{head}_{1}, 
 \dots, \mathit{head}_{h})  \textbf{W}^{O},
\end{split}
\end{equation}
An advantage of using multiple attention heads is that it allows the model to learn to attend over different positions in the input image per attention head during each step of the encoding process. 
For self-attention during encoding, the matrices $\textbf{Q}$,  $\textbf{K}$, $\textbf{V}$ are consistent per layer (i.e., $\textbf{Q} = \textbf{K} =\textbf{V})$ and obtained from the output of the previous layer. In the first layer, they can be obtained from the \ac{VFEN}.

The weights of each transformer layer are not shared between encoder layers. 
We apply the positional encoding as introduced in \citep{vaswani2017attention}. 

\subsection{Character decoding}

The decoder consists of $n$ transformer decoder layers. 
For both the encoder and the decoder we use a transformer architecture. 
The reason to choose a transformer over an RNN-based architecture is that an RNN already has an inductive bias in terms of decoding and encoding direction due to the recurrent nature of the architecture. 
The decoder takes the embeddings of the decoded output character sequence as input. 
Each decoder layer consists of three sublayers: two multi-head attention layers and one feed-forward neural network (the same implementation as in Section~\ref{sec:encoder}). 
The first multi-head attention layer attends over the decoded output characters (\textit{decoder self-attention}). 
The second layer of multi-head attention (\textit{decoder cross-attention}) attends over the encoded \ac{VFE}s from the last encoder layer.  
The decoder cross-attention is able to look at the encoded input image at every step during decoding.
This makes it possible to attend over different encoded image regions during decoding. 
Previous work on \ac{STR} \citep{shi2018aster,cheng2017focusing, zhan2018esir} only uses one attention distribution over the encoded states per decoding step. 
In contrast, per decoding layer $n$, we have $h$ attention heads modeling complex alignments between encoder features and decoded output characters. 

We add an extra direction embedding in order to add more context information by using additional embeddings. This direction embedding tells the model to decode the output sequence from left-to-right or from right-to-left. By adding the direction embedding, we can use the same decoder network and still condition on the output sequence reading direction.  

For  every decoding step $t$, the output embedding $\textbf{h}_{n}$ of the stack of transformer decoders is passed through a feed-forward layer with the output characters as the output space. 
A softmax is applied to obtain a distribution over all output characters. 
During training, this results in a $ V \times L $ matrix, where $V$ is the size of the output character space (or vocabulary) and $L$ the length of the predicted character sequence. 

\subsection{Direction embedding}

We define the decoding direction of the output sequence as two sub-tasks of \ac{STR}. Each decoding direction is one sub-task of the method on which we condition the output sequence. 
To condition the output on a decoding direction, we randomly initialize two 512-d vectors at the start of training. 
During each training iteration, every input image in the batch is decoded twice; once from left-to-right and right-to-left. 
The ground truth description of the right-to-left decoded character sequence is just the reserved ground truth of the original description. 
During decoding, we add the direction embedding on top of the positional embedding and the token embedding. 
This is another way to provide additional context information to the model, similar to the position embeddings. 
Based on this information, the model should learn to decode the character sequence not only in the left-to-right direction but also in the other direction, otherwise the loss function for the right-to-left decoded images will not be minimized.

Similar to the character embeddings, the direction embeddings are trained end-to-end with the rest of the model.

\section{EXPERIMENTAL SETUP}
\subsection{Datasets}\label{sec:dataset}

\ac{Bi-STET} is trained on two synthetically generated datasets. After training, the method is evaluated on seven real-word evaluation sets which are commonly used for scene text recognition.
\subsubsection{Training datasets}
\begin{itemize}[nosep]
\item \emph{Synth90K.} The \textit{Synth90K dataset} \citep{Jaderberg14} is a synthetically generated dataset for text recognition. 
It contains 7.2 million training images.  
The lexicon used contains 90,000 words. 
Each word has been used to render 100 different synthetic images. 

\item \emph{SynthText.} The \textit{SynthText dataset} \citep{gupta16} contains 800,000 synthetically generated images for text detection and recognition, with roughly 8 million annotated text instances placed in natural scenes. We crop all text instances from the original input images, by taking the smallest horizontally aligned bounding box around the annotated text instances in the image. We discard bounding boxes that are smaller than 32 pixels in height or 30 pixels in width. 
Bounding boxes larger than 800 pixels in width, 500 pixels in height or with a transcription label longer than 25 characters are removed too. 
We obtain 2.9 million cropped-word images from this dataset for training.
\end{itemize}

\subsubsection{Evaluation datasets}

\ac{Bi-STET} is evaluated cross-dataset. The model is trained only on synthetically generated word images while we evaluate on real-world word images. Unless stated otherwise, we use the word-image crops and annotations as provided by the dataset to be consistent with other methods. This might result in over-cropped word images; in other cases adding a margin may lead other artifacts.

\begin{itemize}[nosep]
\item \emph{ICDAR03.} The \textit{ICDAR03 dataset}  \citep{lucas2003icdar} contains 258 images for training and 251 for testing. For the text recognition task only, 1,156 word instances can be cropped from the test set. This dataset was collected for the text detection and recognition task. Therefore, most text instances in the images are clearly horizontally visible and centred in the image \citep{veit2016coco}. Following \citep{shi2018aster, cheng2017focusing, wang2010word, wang2011end}, we ignore all words that are shorter than three characters or contain non-alpha numeric characters during evaluation. 

\item \emph{ICDAR13.} The \textit{ICDAR13 dataset}  \citep{karatzas2013icdar} contains most images from the ICDAR03 dataset. In total this dataset contains 1,095 word images for evaluation. Similar to \cite{shi2018aster}, we add a cropping-margin of 15\% to prevent over-cropping.

\item \emph{ICDAR15.} The \textit{ICDAR15 dataset} \citep{karatzas2015icdar} contains 2,077 word images for evaluating. 
The word images are cropped from video frames collected with the Google Glass device. 
These frame crops contain substantial real-world interference factors such as: occlusions, motion blur,  noise, and illumination factors, which are not present in the ICDAR03 and ICDAR13 datasets. 
Like \citet{cheng2017focusing}, we remove all  examples where the ground truth transcription contains non-alpha numeric characters.

\item \emph{SVT.} The \textit{Street View Text dataset} (SVT) \citep{wang2011end, wang2010word} contains images that have been taken from Google Street View. 
Due to this origin, some images have a low resolution and/or contain distortion factors such as noise or blur. This dataset contains 647 word images for evaluation. Per image, a 50-word lexicon is provided as well. Similar to \cite{shi2018aster}, we add a cropping-margin of 5\% to prevent over-cropping. Similar to \cite{shi2018aster}, we add a cropping-margin of 5\% to prevent over-cropping.

\item \emph{SVTP.} The \textit{Street View Text Perspective dataset} (SVTP) \citep{quy2013recognizing} contains 645 word images cropped from Street View. Most images have perspective distortions due to the camera viewpoint angle. 

\item \emph{IIIT-5K Word.} The \textit{IIIT-5k Word dataset} (IIIT5K) \citep{mishra2012scene} contains 3,000 images for evaluation. The word images are cropped from scene texts and born-digital images. For this dataset, per evaluation image, two lexicons of 50 and 1,000 words are provided for lexicon inference. 

\item \emph{CUTE80.} The \textit{Curved Text dataset} (CUTE80) \citep{risnumawan2014robust} mainly contains curved and/or oriented text instances. The dataset was originally proposed for text detection, but later annotated for text recognition as well. In total, 288 high resolution word images can be cropped from the original dataset. 
\end{itemize}

\subsection{Implementation details}


\begin{table*}[htb!]
	\caption{Accuracy of left-to-right vs.\ right-to-left vs.\ bidirectional word decoding. Measured without lexicon and compared with the method by \citet{shi2018aster}. }
	\label{table:bidirect}
	\setlength{\tabcolsep}{3.25pt}
	\centering
		\begin{tabular}{l@{~}ccccccc}
			\toprule
			\bf Method & \bf IIIT5k & \bf SVT & \bf IC03 & \bf IC13 & \bf IC15 & \bf SVTP & \bf CUTE \\
			\midrule 
			 \citet{shi2018aster}, left-to-right      & 91.93                 & 88.76               &  93.49	           & 89.75              &--                  & 74.11  & 73.26 \\
		      \citet{shi2018aster}, right-to-left     & 91.43                 & 89.96               &  92.79             & 89.95              & --                 & 73.95 & 74.31 \\
			 \citet{shi2018aster},   bidirectional    &   92.27               & \textbf{89.5}                & 93.60              & 90.54              &  --                & 74.26 &  74.31 \\
			\midrule 
		      \ac{Bi-STET} (this paper), left-to-right  & 94.2                 & 88.3               & 95.1               & 92.5                & 75.0               & 78.8  & 81.8 \\
		       \ac{Bi-STET} (this paper),  right-to-left & 94.1                 & 87.9               & 95.3               & \textbf{93.4}               & 73.2                 & 79.5 & \textbf{83.6}\\
			  \ac{Bi-STET} (this paper),  bidirectional & \textbf{94.7}   & 89.0 & \textbf{96.0} & \textbf{93.4} & \textbf{75.7} & \textbf{80.6} & 82.5 \\
			\bottomrule
		\end{tabular}
	\label{tbl:bidiretional}
\end{table*}

Our implementation consists of a feature extraction network followed by an encoder-decoder network. 
The code and checkpoint files used to train and evaluate Bi-STET are available at \url{https: //github.com/MauritsBleeker/Bi-STET}.

\subsubsection{Feature extraction network}
All input images are resized to $32 \times 256$ without keeping the original aspect ratio. 
The maximum output sequence length during training is 24. All pixels are normalized with a per-channel calculated mean and standard deviation calculated on the ImageNet dataset \citep{deng2009imagenet}.

\subsubsection{Encoder-decoder network}
For the encoder and decoder we use exactly the same configuration as the base model described in \citep{vaswani2017attention}. We use a stack of $n=6$ transformer layers for both the encoder and decoder. Each layer has eight attentions heads $(h=8)$. The  embedding dimensionality is set to $d = 512$. For the hidden state of the two layer feed-forward network in each transformer layer, we set $d_{f} = 2048$. 

The output space of our model contains all the lower-case characters $\{ a, \dots, z \}$, digits $\{ 0, \dots, 9 \}$, 32 ASCII punctuation marks, similar to \citep{cheng2017focusing, shi2018aster, zhan2018esir}, and a start- and end-of-word symbol. The punctuation marks are included during training, but ignored during evaluation. All evaluation and training ground truth descriptions are lower-case, which makes the model case-insensitive. 

\subsection{Optimization}

The entire method is trained from scratch. All the weights are initialized with Xavier initialization \citep{glorot2010understanding}. 
Similar to \citep{zhan2018esir, shi2018aster, cheng2017focusing, shi2017end, shi2016robust, yang2017learning}, we use ADADELTA  \citep{zeiler2012adadelta} as the optimizer for the model. ADADELTA has a self-adaptable learning rate, which we initialize to 1 
Even though the learning rate of ADADELTA is self-adaptable, we apply a learning rate schedule where we reduce the initial learning rate by a factor of 0.1 after 150,000, 300,000 and 400,000 training iterations. 
Similar to  \citep{shi2018aster}, we find that a  learning rate schedule is beneficial to the performance.

The model is trained for 500,000 training iterations in total, after which it converges. We use Kullback-Leibler divergence as the loss function. 
The batch size is set to 64. 
For each training batch we sample 32 images from the Synth90k dataset and 32 from the SynthText. 
\citet{shi2018aster} and \citet{zhan2018esir} show that methods optimized with balanced batch (of size 64) on the SynthText and Synth90k datasets outperform methods solely trained on Synth90k. Per forward-backward pass, we decode the characters per example from left-to-right and from right-to-left. 

During training, we do one forward pass for left-to-right decoding and one for right-to-left and accumulate the gradients. 
It is possible to train both decoding directions with one forward pass, but for computational reasons we have chosen gradient accumulation instead. 

\subsection{Metrics}

We use the same evaluation metrics as in \citep{cheng2017focusing, shi2018aster, zhan2018esir}. 
The text recognition task includes 68 characters in total. 
During evaluation the 32 ASCII punctuation marks are ignored.  
When a lexicon is provided, the word from the lexicon with the shortest edit distance is selected as the prediction. 
Only predicted sequences of characters that are completely correct are considered to be correctly predicted examples.  
We select the character with the highest probability per index in the sequence, until the end-of-word character is predicted. 
When decoding bidirectionally, the sequence with the highest product probability is selected as final output sequence.

\section{RESULTS}

\begin{table*}[htb!]
\caption{Accuracy compared to state-of-the-art. ST is short for the SynthText dataset, 90K for the Synth90K dataset; 50, 1k, full and 0 are the size of the used lexicons; 0 means that no lexicon is used.}
\label{table:sota-recognition}
\centering
\resizebox{\textwidth}{!}{
\begin{tabular}{llcccccccccccc}
\toprule
\multirow{2}{*}{\bf Method} & \multirow{2}{*}{\bf ConvNet, Data} & \multicolumn{3}{c}{\bf IIIT5k} & \multicolumn{2}{c}{\bf SVT} & \multicolumn{3}{c}{\bf IC03} & \bf IC13 & \bf IC15 & \bf SVTP & \bf CUTE\\
\cmidrule(r){3-5} \cmidrule(r){6-7} \cmidrule(r){8-10} \cmidrule(r){11-11} \cmidrule(r){12-12} \cmidrule(r){13-13} \cmidrule(r){14-14} 
 &  & 50 & 1k & 0 & 50 & 0 & 50 & Full & 0 & 0 & 0 & 0 & 0 \\
\midrule
\citet{su2014accurate}    & -- & -- & -- & -- & 83.0 & -- & 92.0 & 82.0 & -- & -- & -- & -- & --\\
\citet{jaderberg2016reading}  & VGG, 90k & 97.1 & 92.7 & -- & 95.4 & 80.7 & 98.7 & 98.6 & 93.1 & 90.8 & -- & -- & --\\
\citet{jaderberg2014deep}  & VGG, 90k & 95.5 & 89.6 & -- & 93.2 & 71.7 & 97.8 & 97.0 & 89.6 & 81.8 & -- & -- & --\\
\citet{shi2017end}& VGG, 90k & 97.8 & 95.0 & 81.2 & 97.5 & 82.7 & 98.7 & 98.0 & 91.9 & 89.6 & -- & -- & --\\
\citet{shi2016robust}  & VGG, 90k & 96.2 & 93.8 & 81.9 & 95.5 & 81.9 & 98.3 & 96.2 & 90.1 & 88.6 & -- & 71.8 & 59.2\\
\citet{lee2016recursive} & VGG, 90k & 96.8 & 94.4 & 78.4 & 96.3 & 80.7 & 97.9 & 97.0 & 88.7 & 90.0 & -- & -- & --\\
\citet{yang2017learning} & VGG, Private & 97.8 & 96.1 & -- & 95.2 & -- & 97.7 & -- & -- & -- & -- & 75.8 & 69.3\\
\citet{cheng2017focusing} & ResNet, 90k+ST$^{+}$ & 99.3 & 97.5 & 87.4 & 97.1 & 85.9 & \textbf{99.2} & 97.3 & 94.2 &93.3 & 70.6 & -- & --\\
\citet{shi2018aster}  & ResNet, 90k+ST & \textbf{99.6} & 98.8 & 93.4 & 97.4 & 89.5 & 98.8 & 98.0 & 94.5 & 91.8 & 76.1 & 78.5 & 79.5\\
\citet{zhan2018esir} & ResNet, 90k + ST & \textbf{99.6} & 98.8 & 93.3 & 97.4& \textbf{90.2}& - & - & - & 91.3 & 76.9 &79.6 & 83.3 \\
\citet{sheng2018nrtr} & Modality-Transform, 90k & 99.2 & 98.8 & 86.5 & \textbf{98.0} & 88.3 & 98.9 & 97.9 & 95.4 & \textbf{94.7} & - & - & - \\
\citet{yang2019symmetry} & ResNet, 90k+ST  &  99.5 & 98.8 & 94.4 &  97.2 & 88.9 & 99.0 & 98.3 & 95.0 & 93.9 & \textbf{78.7} & \textbf{80.8} & \textbf{87.5} \\
\midrule
Bi-STET (this paper) & ResNet, 90k+ST & \textbf{99.6} & \textbf{98.9} & \textbf{94.7} & 97.4 & 89.0 & 99.1 & \textbf{98.7} & \textbf{96.0} & 93.4 & 75.7 & 80.6 & 82.5 \\
\bottomrule
\end{tabular}%
}
\end{table*}

First, we compare the bidirectional sequence predicting for \ac{STR} with a single decoder vs.\ with two decoders. 
Next,  we examine the performance of \ac{Bi-STET} and other models on \ac{STR} evaluation sets.
Finally, we provide analyzes of the attention mechanism in \ac{Bi-STET} and the capability of the method to handle curved and rotated text.

\subsection{Bidirectional decoding}
\begin{table}
\label{table:ota-recognition}
\centering
\caption{Comparison of the number of trainable model parameters and training iterations.}
\begin{tabular}{lccc}
\toprule
\bf Method    & \bf Model  & \bf Training  & \bf Batch \\ 
    & \bf parameters  & \bf  iterations  & \bf size \\ 
 & ($ \times 10^6$) & ($\times 10^6$) &  \\
\midrule
\citet{shi2018aster} &  88                                                 & 1\phantom{.125}            & 64      \\
\midrule
\ac{Bi-STET} (this paper)      &\textbf{66}                                                 & \textbf{0.5}\phantom{12}                   & 64                                           \\ 
\bottomrule
\end{tabular}
\label{table:params}
\end{table}

Similar to \citep{shi2018aster}, we condition the output character sequence on a decoding direction. We validate our universal bidirectional decoding with three evaluation variants, similar to \citet{shi2018aster}. 
For the first variant, we decode the output sequence from left-to-right, by only using the left-to-right direction embedding. 
In the second variant, we only use the right-to-left directional embedding. 
In the third variant, we decode each evaluation example twice, once with each direction embedding. 
The two predicted outputs can have different sequence lengths. For each prediction (left-to-right and right-to-left) we take the sequence with the highest probability by taking the arg-max for each position and take the product of the probabilities as the probability of the entire sequence. 
We select the sequence with the highest output probability as the final prediction. In case that the right-to-left prediction has the highest probability, we reverse the sequence to match with the ground truth. 

In Table~\ref{table:bidirect} we show the results of the three aforementioned evaluation variants and the results obtained by \citet{shi2018aster}. 
For 6 out of 7 evaluation sets, we achieve state-of-the-art results for bidirectional \ac{STR}. 
For 6 out of 7 of the evaluation sets, \ac{Bi-STET}s bidirectional decoding leads to higher scoring sequence prediction than using a single decoding direction. 
Only for the CUTE80 set, the right-to-left decoding leads to a higher accuracy score than bidirectional. 
The gain in performance due to the bidirectional decoding is similar as in the method by \citet{shi2018aster}.  

We also show  that, by using a transformer-based encoder-decoder, bidirectional \ac{STR} can be substantially simplified in comparison to the method by \cite{shi2018aster}. 
In Table \ref{table:params}, we compare the number of model parameters and the number of training iterations with the method by \citet{shi2018aster}. 
We use similar training settings, in terms of batch size, optimization, data, etc. as \citep{shi2018aster}.
Based on Table~\ref{table:params}, it is clear that with a single bidirectional transformer decoder, the number of training iterations can be reduced by 50\% compared to methods that use two separate RNN decoders -- in combination with significantly fewer model parameters. 
Fewer training iterations and model parameters are excellent properties from an efficiency and computational point of view. We also outperform the RNN based method, which also uses an extra image rectification network, on most evaluation sets.

To summarize, the results in Table~\ref{table:bidirect}  and Table \ref{table:params} show that similar or better results can be obtained with significant less training parameters and twice the efficiency by using a single transformer decoder.

\subsection{Text recognition}

In Table \ref{table:sota-recognition}, we evaluate \ac{Bi-STET}  in terms of prediction accuracy on 7 public evaluation sets and compare it to other \acf{SOTA} \ac{STR} methods. \ac{Bi-STET} meets or outperforms \ac{SOTA} methods on 6 out of 12 evaluation experiments. We achieve new \ac{SOTA} results on the ICDAR03 and the IIIT5K datasets. 

The strength of the transformer encoder-decoder w.r.t. to images with oriented and curved text is also shown by the results in  Table \ref{table:sota-recognition}. 
The five datasets where STET does not beat, but meets, the state-of-the-art are CUTE80, ICDAR13, ICDAR15, SVT-P and SVT. 
The fact that we do not achieve the state-of-the-art on the datasets SVT-P, ICDAR15 and CUTE80 can be explained by the fact that those datasets contain a considerable number of images that are rotated or have perspective distortions. 
We meet \ac{SOTA} results on these datasets, we speculate that we don't exceed them because they have these distortions. It should also be noted we are able to meet these \ac{SOTA} results without any specific network component for dealing with distortions, which other methods explicitly require \cite{yang2019symmetry, zhan2018esir, shi2018aster}.
For ICDAR13 and SVT-P we do not establish new \ac{SOTA} performance figures, although we do meet the results of other methods with a small margin. 

\subsection{Analyzes}
\subsubsection{Attention heads analyzes}
\begin{figure*}[htb!]
\centering
\begin{subfigure}[b]{0.8\textwidth}
   \includegraphics[clip,trim=0mm 50mm 0mm 40mm,width=1.0\textwidth]{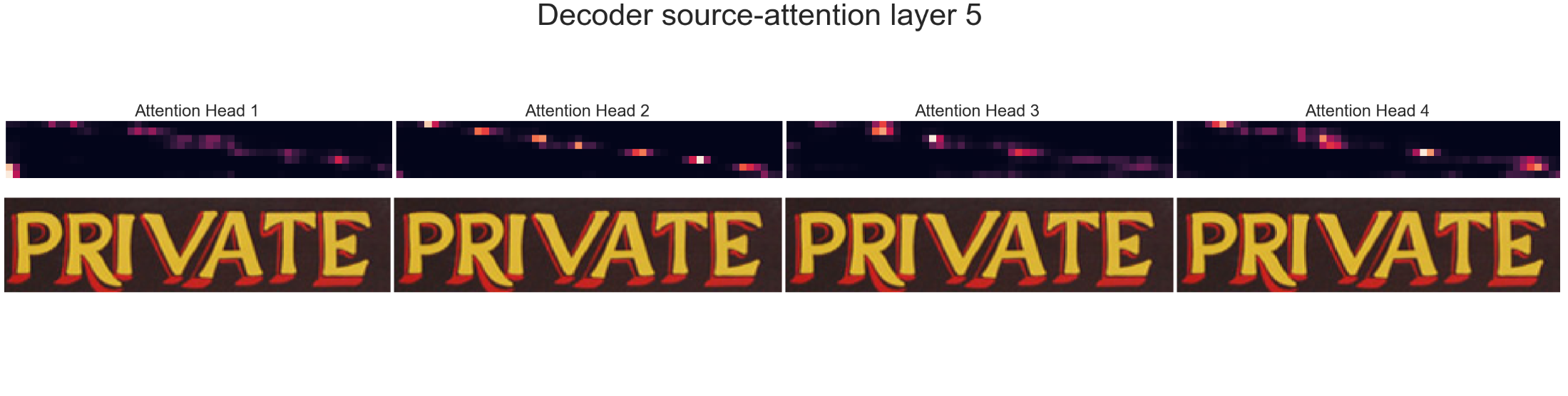}
   \caption{Decoding attention while decoding from left-to-right.}
   \label{fig:encoder-self-attention-ltr} 
\end{subfigure}
\begin{subfigure}[b]{0.8\textwidth}
   \includegraphics[clip,trim=0mm 50mm 0mm 40mm,width=1.0\textwidth]{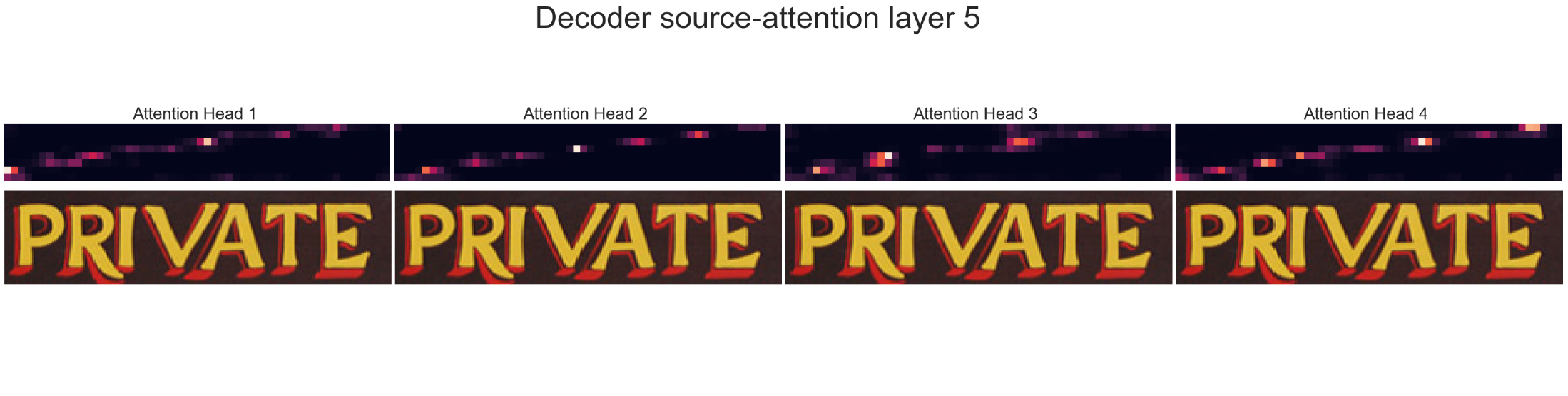}
   \caption{Decoding attention while decoding from right-to-left.}
   \label{fig:encoder-self-attention-rtl}
\end{subfigure}
\caption[]{Visualization of the self-attention of layer 5 of the decoder while decoding the sequence (left-to-right). Each row in the matrix visualizes the attention distribution over the embeddings of the image on the X axis, while decoding the corresponding character in the input.}
\end{figure*}

To get an understanding of the internal behaviour of \ac{Bi-STET}, we extracted the attention distributions from \ac{Bi-STET} during evaluation and visualize them in Figures~\ref{fig:encoder-self-attention-ltr} and \ref{fig:encoder-self-attention-rtl}.
Conditioned on different decoding directions, \ac{Bi-STET} has learned to model an inverse alignment between the predicted output character and the region in the input images where the character is depicted -- using only a single decoder. 
In Figure~\ref{fig:encoder-self-attention-ltr}, there is a clear attention alignment going from left to right over the image, while in Figure~\ref{fig:encoder-self-attention-rtl}, this alignment goes in the opposite direction. 
The model has jointly learned to model the character-image region alignment in both directions. 
Also, different attention heads do not specialize for left-to-right or right-to-left decoding, but learn how to change the attention direction when the output is conditioned on a different decoding direction. This shows the strength the method w.r.t. regularisation towards both sub-tasks.
This is interesting from a multi-task learning point of view because this indicates that the same attention heads can learn different (sub)tasks. 

\subsubsection{Rotated and curved text}

\begin{figure}[tb!]
    \centering
    \begin{subfigure}[t]{0.24\columnwidth}
        \includegraphics[width=\columnwidth]{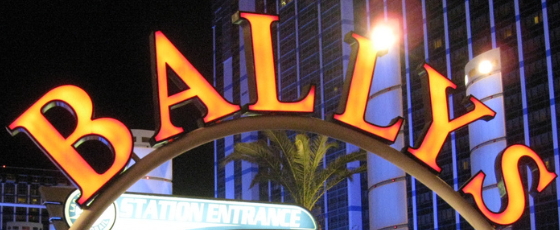}
       \caption{\raggedright\ ballys vs. \textcolor{red}{bally}}
   \end{subfigure}
 \begin{subfigure}[t]{0.24\columnwidth}
        \includegraphics[width=\columnwidth]{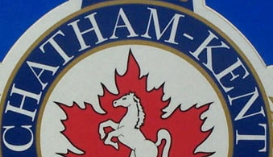}
       \caption{\raggedright chathamkent vs. \textcolor{red}{from}}
    \end{subfigure}
 \begin{subfigure}[t]{0.24\columnwidth}
        \includegraphics[width=\columnwidth]{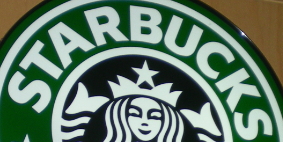}
         \caption{\raggedright starbucks vs. \textcolor{red}{and}}
    \end{subfigure}
 \begin{subfigure}[t]{0.24\columnwidth}
        \includegraphics[width=\columnwidth]{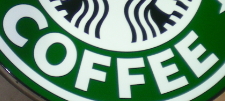}
        \caption{\raggedright coffee vs. \textcolor{red}{offer}}
    \end{subfigure}
\begin{subfigure}[t]{0.24\columnwidth}
        \includegraphics[width=\columnwidth]{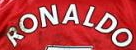}
        \caption{\raggedright  ronaldo vs. \textcolor{ForestGreen}{ronaldo}}
    \end{subfigure}
 \begin{subfigure}[t]{0.24\columnwidth}
        \includegraphics[width=\columnwidth]{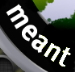}
        \caption{\raggedright meant vs. \textcolor{ForestGreen}{meant}}
    \end{subfigure}
 \begin{subfigure}[t]{0.24\columnwidth}
        \includegraphics[width=\columnwidth]{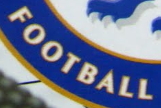}
        \caption{\raggedright  football vs. \textcolor{ForestGreen}{football}}
    \end{subfigure}
 \begin{subfigure}[t]{0.24\columnwidth}
      \includegraphics[width=\columnwidth]{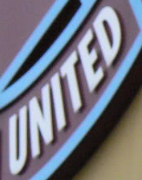}
        \caption{\raggedright  united vs. \textcolor{ForestGreen}{united}}
    \end{subfigure}
    \caption{Examples of curved text examples from the CUTE80 dataset that are \textcolor{ForestGreen}{correctly} and \textcolor{red}{incorrectly} predicted by \ac{Bi-STET}. In black the ground truth is given.}
    \label{fig:curved-words}
\end{figure}

\ac{Bi-STET} is solely trained as a general image-to-text encoder-decoder and does not contain a specific rectification component for handling rotated or curved text instances, unlike previous methods ~\citep{shi2018aster, zhan2018esir, yang2019symmetry}. 
We are able to obtain results that meet those of state-of-the-art methods that are specifically optimized for curved and rotated text. 
Figure~\ref{fig:curved-words} provides a sample from the CUTE80 dataset with correctly and incorrectly predicted sequences. 
Looking at correctly predicted examples, we see that \ac{Bi-STET} properly decodes words that are slightly curved or only curved in one direction.  This is where the bidirectional decoding shows its strength. 
For example, the two middle images of the second row are correctly decoded when decoding from right-to-left decoder, but not when decoding in the other direction.
From the first row of images we see that words that are curved in very strong arc shapes (heavy perspective distortions) are difficult for \ac{Bi-STET} to decode. This also shows the strength of method w.r.t. regularisation towards curved and rotated text without using any specific rectification component.

\subsubsection{Sequence length}
\begin{figure}[htb!]
	\centering
	\includegraphics[width=0.9\columnwidth]{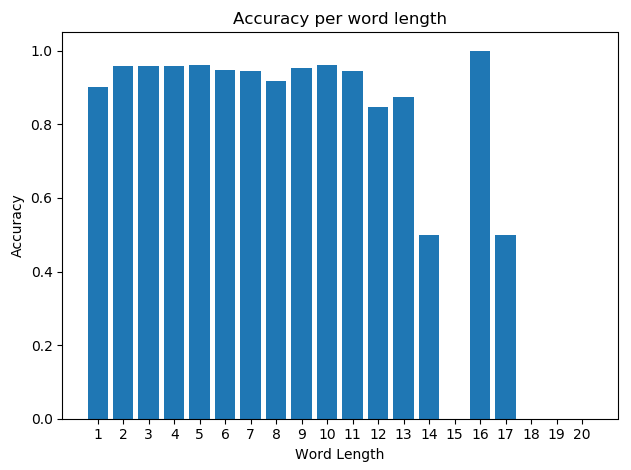}
	\caption{Text Recognition accuracies versus word length for \ac{Bi-STET}. Tested on IIIT-5K.}
 	\label{fig:word-accuracy}
\end{figure}

\citet{vaswani2017attention} argue that transformer-based architectures are more suitable for capturing long-range dependencies for machine translation than \ac{RNN}s, because of the global attention per encoding and decoding step. 
The self-attention results in the fact that the maximum distance in a sequence between two embeddings which are encoded or decoded is 1. 
Despite the fact that the maximum output sequence length in our evaluation experiment (max. length 17) is not as long as for other language tasks~\cite{khandelwal2018sharp}, we are interested in whether or not our transformer-based method is better in predicting longer output character sequence than an \ac{RNN}-based method. 
In Figure~\ref{fig:word-accuracy}, we show the relation between output sequence length and accuracy for the IIIT-5K evaluation set. 
We can see that the prediction accuracy given a sequence length is more or less constant until we reach a character length of 11. 
After a sequence length of 11, the accuracy starts to degrade. 
It should be noted that there are very few samples with a sequence length of 11 or higher. 

By comparing Figure~\ref{fig:word-accuracy} to Figure~12 in \citep{shi2018aster}, we see that \ac{Bi-STET} performs similarly to \citet{shi2018aster}'s method for short character sequences and slightly better for longer character sequences which are longer than 11 characters. 
We conclude that \ac{STET} performs similar for short characters sequences and at least as good in decoding longer character sequences. 

\section{DISCUSSION AND CONCLUSION}

We have introduced \ac{Bi-STET}, a method for bidirectional \ac{STR} with a single decoder.
\ac{Bi-STET} is capable of bidirectional decoding, without implementing the decoding direction conditioning at the architecture or algorithm level.  
The decoding conditioning is implemented at the input level, by adding an extra direction embedding to the input.  

We show that \ac{Bi-STET} achieves or outperforms state-of-the-art \ac{STR} methods, with a considerably more efficient approach than other bidirectional \ac{STR} methods (i.e., requiring 50 \% less training iterations and significant less model parameters). 
By having fewer model parameters, the model can be executed on devices with less computational resources (for user applications). Besides that, less computational resources are required to obtain \ac{SOTA} text recognition results. 
We also show that \ac{Bi-STET} learns to exploit the same attention heads for both decoding directions, which means that there are no specialized attention heads in the model for each decoding direction. 
This is interesting from a multi-task learning point of view because different heads tend not to be focused on one decoding direction. 
Finally, we show that, due to the bidirectional decoding, \ac{Bi-STET} is capable of handling slightly curved and orientated text and performs as well for longer text sequence as other bidirectional \ac{STR} methods.

A future research direction is to combine \ac{Bi-STET} with a Spatial Transformer Network~\citep{shi2016robust, shi2018aster, yang2019symmetry} or a Rectification Network~\citep{zhan2018esir}. 
\ac{Bi-STET} is able to handle oriented and perspective text in images; we believe that \ac{Bi-STET} could benefit from an extra image processing component to be able to better handle oriented or perspective text. 
From a multi-task learning point of view, it would be interesting to explore task conditioning on the input level with more diverse tasks. 
In addition, an extension to tasks with more complex and diverse data modalities would also be a possible future research direction.

\ack We thank Ana Lucic, Maartje ter Hoeved and Mostafa Dehghani for comments and suggestions. This research was supported by Ahold Delhaize, the Association of Universities in the Netherlands (VSNU), the Innovation Center for Artificial Intelligence (ICAI), and the Nationale Politie. All content represents the opinion of the authors, which is not necessarily shared or endorsed by their respective employers and/or sponsors.

\bibliographystyle{abbrvnat}
\bibliography{references}

\end{document}